\documentclass[11pt]{article}

\usepackage{blindtext}

\usepackage[preprint]{jmlr2e}
\usepackage[numbers,sort&compress]{natbib}
\usepackage{graphicx}
\usepackage{tikz}
\usepackage{placeins}
\usepackage{makecell}
\usepackage{subfigure}
\usepackage{amsfonts}       
\usepackage{amsmath,amssymb}
\usepackage{pgfplots}
\pgfplotsset{compat=1.18}
\usepackage{bbm}
\usepackage{empheq}
\usepackage{xargs}
\usepackage{cleveref}

\DeclareMathOperator*{\argmax}{arg\,max}

\usepackage{lastpage}

\firstpageno{1}

\begin{document}

\title{Stepping Out of the Shadows: Reinforcement Learning in Shadow Mode}

\author{\name Philipp Gassert\hspace{1pt}\thanks{Correspondence: \texttt{philipp.gassert@tum.de}} \hspace{2pt}\affnum{1,2}
       \and
       \hspace{2pt}
       \name Matthias Althoff \affnum{1,2}\\
        [6pt]
       \affiliation{1}{Technical University of Munich} \\
       \affiliation{2}{Munich Center for Machine Learning (MCML)}
}

\maketitle

\begin{abstract}
Reinforcement learning (RL) is not yet competitive for many cyber-physical systems, such as robotics, process automation, and power systems, as training on a system with physical components cannot be accelerated, and simulation models do not exist or suffer from a large simulation-to-reality gap.
During the long training time, expensive equipment cannot be used and might even be damaged due to inappropriate actions of the reinforcement learning agent.
Our novel approach addresses exactly this problem: We train the reinforcement agent in a so-called shadow mode with the assistance of an existing conventional controller, which does not have to be trained and instantaneously performs reasonably well.
In shadow mode, the agent relies on the controller to provide action samples and guidance towards favourable states to learn the task, while simultaneously estimating for which states the learned agent will receive a higher reward than the conventional controller.
The RL agent will then control the system for these states and all other regions remain under the control of the existing controller.
Over time, the RL agent will take over for an increasing amount of states, while leaving control to the baseline, where it cannot surpass its performance.
Thus, we keep regret during training low and improve the performance compared to only using conventional controllers or reinforcement learning.
We present and evaluate two mechanisms for deciding whether to use the RL agent or the conventional controller. 
The usefulness of our approach is demonstrated for a reach-avoid task, for which we are able to effectively train an agent, where standard approaches fail.
\end{abstract}



\section{Introduction} \label{sec:introduction}

Reinforcement learning (RL) has made impressive progress in recent years and is rather easy to implement wherever simulations of a given environment are available~\citep{mnih2015human, Silver_Schrittwieser_Simonyan_Antonoglou_Huang_Guez_Hubert_Baker_Lai_Bolton_2017, Wurman_Barrett_Kawamoto_MacGlashan_Subramanian_Walsh_Capobianco_Devlic_Eckert_Fuchs_2022}. 
RL bears great potential for cyber-physical systems, but is not yet competitive in many domains. 
Obtaining a high fidelity simulation environment and transferring policies trained in simulation to the physical system proves difficult, but training on the physical system is often unreasonable, due to time and resource constraints and risks involved with running randomly initialized RL agents~\citep{Rajeswaran_Kumar_Gupta_Vezzani_Schulman_Todorov_Levine_2017, Zhao_Queralta_Westerlund_2020, Wang_Bao_Clavera_Hoang_Wen_Langlois_Zhang_Zhang_Abbeel_Ba_2019, Zhu_Guo_Owaki_Kutsuzawa_Hayashibe_2023}.
The cost inferred by the ineffective operation of the system during training can be prohibitive, especially whenever an existing control law is available~\citep{Sünderhauf_Brock_Scheirer_Hadsell_Fox_Leitner_Upcroft_Abbeel_Burgard_Milford_etal._2018, Panaganti_Xu_Kalathil_Ghavamzadeh_2022}.
Offline learning methods, such as behavioural cloning and offline RL~\citep{Kumar_Fu_Tucker_Levine_2019, Kanervisto_Karttunen}, address this issue by using static data provided by expert demonstrations.
Offline methods cannot explore, and are dependent on the quality of the expert trajectories and the coverage of the state space by the expert~\citep{Levine_Kumar_Tucker_Fu_2020}.

In this work, we propose training an RL agent on the system, without giving it sole control, which we call training in shadow mode.
We assume that an existing but suboptimal policy is available or can easily be implemented, e.g.\ from a conventional controller or human operator.
For example, we might be able to craft a basic heuristic for training in shadow mode that exploits some task knowledge, as a starting solution for our RL agent.
We refer to this existing policy as the baseline, since we assume that it performs well, meaning it yields an average reward substantially above that of a random policy. 
At each time step, our novel algorithm chooses between the actions provided by the RL agent and the existing controller.
Training in shadow mode thus focuses on a type of hybrid RL where baseline and RL agent actions are not mixed as a weighted sum, but combined by switching between their policies. 
The RL agent trains ``in the shadow'', using actions proposed by the controller for training, and taking control for increasingly many states.
When the actions of the RL agent are chosen, it can explore beyond the state distribution given in trajectories created by the baseline alone.
The agent learns its policy and simultaneously identifies states where it yields a higher reward than the baseline.
For these states, the RL agent controls the system, otherwise, control is left to the baseline.
The resulting combination of baseline and RL agent creates a combined agent with an expected reward above the individual expected rewards of the baseline or the RL agent alone.
This includes the possibility of training an RL agent that is always superior to the baseline, but the combined agent is specifically designed to switch back and forth between the two.
Since we train the combined agent to receive higher rewards than the RL agent or the baseline alone, we keep this combination for deployment.
Thus, we can not only exploit the strengths of both agents, but also do not need the RL agent to optimize its policy for the entire state space.
Even though we do not pretrain the agent, we benefit from guided exploration towards favourable states when following the baseline, which is especially helpful in scenarios with sparse rewards~\citep{uchendu2023jump}\@.
Furthermore, we can keep training the agent continuously.

Our main contributions are:
\vspace{-0.3em}    
\begin{itemize}
    \item We propose the shadow mode as a novel training paradigm for RL agents, switching between a preexisting baseline and the RL agent to leverage training on real environments. 
    \vspace{-0.6em}    
\item We propose switching mechanisms for RL agents to take over control based on the $Q$-function or an additional agent action.
\vspace{-0.6em}        
\item We demonstrate our method for a reach-avoid task.
\end{itemize}
\vspace{-0.3em}

The remainder of this paper is structured as follows:
Sec.~\ref{sec:related_work} provides an overview over neighbouring methods in RL\@.
Sec.~\ref{sec:preliminaries} presents the necessary preliminaries.
In Sec.~\ref{sec:rl_in_shadow_mode} we present the proposed algorithm and its different components.
We demonstrate our novel approach using a reach-avoid task in Sec.~\ref{sec:experiments}, and conclude in Sec.~\ref{sec:conclusions}.


\section{Related work} \label{sec:related_work}


\paragraph{Hybrid RL}
There are several approaches to combining RL strategies with control priors.
They typically combine policies as a weighted sum of their actions, where the weighting can either be fixed or adapted during training \citep{Johannink2018ResidualRL, Hoel2020ReinforcementLW, daoudi2023enhancing}.
In contrast, when training in shadow mode, we do not blend policies but choose one or the other at each time step.
\citet{cramer2024contextualized} provide a general formulation for mixing or combining strategies, but also focus on mixing by a weighted sum in their work.

\paragraph{Imitation learning}
Imitation learning refers to methods of inferring a policy from examples or demonstrations of a given task.
The demonstrations are either directly mimicked through behavioral cloning in a supervised learning fashion or used to infer the reward function, in some implementations querying an expert policy for its proposed action~\citep{ross2011reduction,abbeel2004apprenticeship,finn2016guided}. 
There is a number of approaches using imitation learning to pretrain agents, facilitating the (early) training of an otherwise standard RL method~\citep{NIPS1996_68d13cf2, Rajeswaran_Kumar_Gupta_Vezzani_Schulman_Todorov_Levine_2017} and later refining the agent through online training.
In contrast to these approaches, our training method does not explicitly use imitation of the baseline or expert, and thus avoids constraining policies to be close to the expert policy, to achieve better rewards. 

Reinforcement learning in shadow mode benefits from guided exploration similarly to jump-start RL~\citep{uchendu2023jump}.
However, jump-start RL does not consider switching back and forth between baseline and agent, but uses a single switch from guiding example to the RL agent per episode to achieve faster training convergence.
Furthermore, this single switch from guiding example to RL agent is purely schedule-based.

\paragraph{Transfer Learning}
Transfer learning is related to our approach in that it aims at adapting an existing policy~\citep{Zhu_Lin_Jain_Zhou_2023}, but is typically applied with the goal of transferring a policy to a new domain or making the policy more robust to environment changes~\citep{Huang_Feng_Lu_Magliacane_Zhang_2022}.
Training in shadow mode, on the other hand, aims at improving upon an existing controller within one non-varying domain.
\citet{Zhang2023PEX} use a similar approach for combining an offline pretrained agent and an agent that is trained online for offline-to-online transfer. Our method is not aimed at offline-to-online transfer and does not require a pretrained RL agent.

%
%

 \paragraph{Hierarchical Learning}
In hierarchical reinforcement learning, there are several layers of agents, which complete subtasks of the overall problem~\citep{vezhnevets2017feudal, bacon2016optioncritic}.
Parent agents choose from a number of available agents on the next lower level that execute their actions (possibly invoking agents one more level down) until some termination condition is fulfilled.
While the concept of choosing agents at different states is related to our approach, we do not use a hierarchical structure with multiple learned agents, reducing the number of potential points of failure.
Also, to the best of our knowledge, there are no approaches designed to combine an expert-like baseline and a single RL agent.



\section{Preliminaries} \label{sec:preliminaries}
In this section we discuss our basic RL setting, the RL agent, and the baseline.

\paragraph{Markov decision processes}\label{par:problem-formulation}
We consider problems that can be formulated as finite-horizon, discrete-time Markov decision processes (MDPs). 
MDPs \ $\mathcal{M} = \langle \mathcal{S, A}, d^0, p, r, T \rangle$ are described by the initial state distribution $d^0$, the time horizon $T$, a set of states $s_t \in \mathcal{S}$ and actions $a_t \in \mathcal{A}$ for every time step ${t = 0, 1, ..., T}$, the corresponding transition function $p(s_{t+1} \vert s_t,a_t)$, and the reward function $r_t = r(s_t,a_t)$.
For this paper we assume a model-free RL approach, meaning the transition function is not explicitly modeled.
The policy function $\mathit{\pi(s)}$ returns the action $a_t$ for a given state $s_t$.
Many algorithms, such as DDPG~\citep{silver2014deterministic} use a neural network to predict $a_t$ given $s_t$, and add noise to this prediction to enhance exploration.
When the agent executes action $a_t$ according to $\pi$, it receives the appropriate reward $\mathit{r_t}$, eventually obtaining state-action sequences $\mathit{\tau = (s_0,a_0,r_0,s_1,a_1,r_1,...,s_T,r_T)}$.
Alternatively, some algorithms only consider single-step tuples $\mathit{(s_t,a_t,r_t,s_{t+1})}$ for training.
The state distribution (also known as state visitation frequency) for the MDP for time step $t$ under some policy $\pi$ is denoted as $d_{\pi}^t$ and the stationary distribution over time as $d_\pi$.
The task performance is defined as the expected cumulative reward $\mathit{J(\pi)}$ under $\pi$:
\begin{equation}
     J(\pi) = \mathbb{E}_{\tau \sim \pi, p}\left[\sum_{t=0}^T r_t \right]\label{eq:standard-problem-setting}.
\end{equation}
We want to find an optimal policy $\mathit{\pi_*}$ maximizing the reward $\pi_* = \argmax_\pi J(\pi)$.
Regret is defined as the difference in accumulated reward during training between the trained policy $\pi$ and the optimal policy $\pi_*$.
Note that we omit the discount factor and do not consider infinite-horizon settings for simplicity.

\paragraph{Deep reinforcement learning}\label{par:reinforcement-learning-agent}
In recent years, numerous RL methods with impressive performance have evolved, such as TRPO~\citep{schulman2015trust}, PPO~\citep{schulman2017proximal}, ACER~\citep{wang2017sample}, and SAC~\citep{haarnoja2018soft}.
We use DDPG~\citep{silver2014deterministic}, an off-policy actor-critic algorithm that approximates the $Q$-function, which we will use to assess the actions of the actor and the baseline.

In deep RL, the policy $\pi$ is modeled by a neural network, parameterized by its parameters $\theta$, which are then updated to maximize the expected reward $\mathit{J(\pi)}$.
%
Actor-critic methods \citep{NIPS1999_6449f44a} pair the policy network with a value network approximating the value function or $Q$-value function
\begin{equation}
    \mathit{Q(s_t,a_t) = r(s_t,a_t) + \mathbb{E}_{\tau \sim \pi^*, p} \left[ \sum_{k=1}^{T-t} r_{t+k}\right]}\label{eq:q-function}
\end{equation}
such that training can be performed with single-step tuples instead of trajectories $\tau$.

Our novel algorithm builds on the actor-critic method DDPG to predict actions for states and provide $Q$-values for state-action pairs.
During training, the actor predicts actions for given states and is trained to maximize their value, while the critic is trained to approximate the $Q$-function with the tuples collected during training.
Other methods that only approximate the $Q$-function and formulate their policy implicitly by choosing the action with the highest value, can be trained similarly with our novel approach.
We refer to the policy of our agent as $\pi^a$.



\paragraph{Baseline policy}\label{par:baseline-policy}
The baseline used in our approach can be a manually scripted policy, a human, or another fully trained agent, obtained by a different learning scheme.
We denote its policy with $\pi^b$, independently of the formulation of the baseline, which can even be based on a different state space.
As mentioned above, one possibility is to use adaptive control to follow a trajectory on which the RL agent can then improve.
It is also possible to obtain a baseline policy from offline methods if the data on which this agent is trained offline does not yield an optimal policy, or use other pretrained agents if continuing to train them is not viable.
The baseline policy $\pi^b$ does not have to be deterministic.

\paragraph{Problem statement}
For training in shadow mode, we consider problems represented by MDP $\mathcal{M}$ for which a baseline with policy $\pi^b$ that provides actions $a^b$ according to $\pi^b$, for the entire state space $\mathcal{S}$ is available.
The goal is to train the policy $\pi^a$ of an RL agent, while the agent automatically assumes control authority over the system, whenever it expects this action to outperform the baseline action $\pi_b$ according to~\eqref{eq:standard-problem-setting}.
We thus want to learn the combined policy $\pi^c_*(\pi^a, \pi^b)$ and RL agent policy $\pi^a$ maximizing the expected cumulative reward $\mathit{J(\pi^{c}_*)}$:
\begin{equation}
     \pi^{c}_* = \argmax_{\pi^c, \pi^a} \ \mathbb{E}_{\tau \sim \pi^c(\pi^a, \pi^b)}\left[\sum_{t=0}^T r_t \right]\label{eq:problem-statement-shadow}
\end{equation}


\section{Reinforcement learning in shadow mode} \label{sec:rl_in_shadow_mode}
In the standard RL setting, the RL agent policy $\mathit{\pi^{{a}}}$ is trained without the use of other existing policies.
However, when a baseline policy $\mathit{\pi^{{b}}}$ is available, which is a viable but sub-optimal policy for the MDP, we can train the RL agent using our novel shadow mode.
We combine the two policies to obtain the combination $\mathit{\pi^{{c}}}$, as depicted in Fig.~\ref{fig:framework}. 
There are two ways to train the RL agent and simultaneously make decisions on which action to choose, which will be outlined in the following section.
\input{figures/framework.tikz}

\vspace{-2em}
\subsection{Action decision}\label{subsec:action-decision}
The performance of the combined policy $\pi^c$ is crucially determined by the algorithm used for deciding which of the available actions is chosen.
Here, we present two approaches for designing $\pi^c$, which rely on information provided by the RL agent itself to determine whether to assume control authority.

\subsubsection{Control authority by agent} \label{subsubsec:agent_decision}
The first option for giving control authority to the agents is to let the RL agent make this decision explicitly.
We augment the action space, such that the agent has to provide an additional action $a_t^{decision}$ containing the decision of the agent.
More specifically, $a_t^{decision}~\in~\left[ 0, 1 \right]$ is the probability of executing the agent action $a_t^a$, resulting in its total action $a_t^{a,total} = (a_t^a, a_t^{decision})$. 
We define a threshold $\eta_{agent} \in [0, 1]$ which the action value has to cross in order for the actual action, provided through the remaining action values, to be executed on the environment:
\begin{equation}
    a^c_t = \begin{cases}
              a^a_t, & \text{if} \ a_t^{decision} > \eta_{agent} \\
                a^b_t, & \text{otherwise}
    \end{cases}\label{eq:agent-decision}
\end{equation}
The combined policy $\pi^c$ can be learned in a straightforward RL manner.
It is possible to regularize the policy such that the agent predicts actions $a_t^a$ close to $a_t^b$ by introducing the penalty term
\begin{equation}
    r_t^{reg} = -\lambda \lVert a_t^a - a_t^b \rVert\label{eq:reg-action}
\end{equation}
to the reward function, with $\lambda \geq 0$ being the regularization strength.
A potential downside is that the RL agent has to provide a good approximation of $\pi^*$ for the entire state space, even for states that can best be handled by the baseline.
We discuss experimental results in Sec.~\ref{sec:experiments}.

\subsubsection{Control authority by Q-value} \label{subsubsec:max_q_val}
As a second option, for architectures where it is available, we propose to use the $Q$-function to choose actions $a_t^c$.
For approaches where the $Q$-value is not available, one has to use the previous method.
The decision is made by comparing $Q(s_t, a_t^a)$ and $Q(s_t, a_t^b)$ and choosing the action with the larger value.
The resulting action decision is:
\begin{equation}
    a^c_t = \begin{cases}
              a^a_t, & \text{if} \ Q(s_t, a_t^a) > Q(s_t, a_t^b) \\
                a^b_t, & \text{otherwise}
    \end{cases}\label{eq:q-val-decision}
\end{equation}
The transition tuples $(s_t,a^c_t,r_t,s_{t+1})$ used for training the $Q$-function, must contain the actually executed action $a^c$, of course, meaning that in typical implementations the agent action must be replaced with the baseline action if the baseline action was executed.
For using the $Q$-values as suggested, we query the baseline action in every step, calculate both $Q$-values using the critic network, and then make a decision according to these values.
For states in which the baseline performs well, it will be difficult for the agent to predict actions with $Q(s_t, a_t^a) > Q(s_t, a_t^b)$.

\subsection{Exploration and trajectory generation} \label{subsec:exploration}


An important aspect of training in shadow mode is that the combined agent does not always train at the initial time step, but trains at different, randomly chosen steps in time.
At the beginning of a new episode, we follow $\pi^b$ for $t^{train}$ steps, with $t^{train}$ being that randomly chosen start of training.
Therefore, the combined agent and thus the RL agent quickly encounter different states of progression, which are presumably favourable states, depending on how well the baseline policy performs.
As presented by \citet{uchendu2023jump}, training the RL agent from later states in demonstrations can greatly facilitate learning, while training RL agents from scratch can be very difficult.
We also leverage this principle in the shadow mode.
Here, we even extend this method of generating favourable initial states for the RL agent by training from different steps in time: The RL agent may hand back control over the system to the baseline and retake control later.
Switching to the RL agent and then back to the baseline provides additional guidance to the RL agent as it encounters actions $a^b_t$ for states $s_t$ that had not been visited previously, meaning the respective state had had a low state density $d_{\pi^b}(s_t)$ under the policy $\pi^b$ alone.

\section{Experiments} \label{sec:experiments}
We demonstrate our novel approach for a 2D reach-avoid task, and evaluate how the agent learns in shadow mode.
We show that the agent quickly learns to reach the goal position and automatically takes control of the system after reaching sufficient confidence.

\subsection{Environment and agents}
The agent has to maneuver in a continuous 10 by 10 unit $(x,y)$-coordinate frame with a square boundary.
We define a goal position at random at the right half ($x>5$) of this area that the agent that is randomly placed on the left half has to maneuver to.
Furthermore, 95 \% of the scenarios include a randomly placed obstacle, represented by a line segment, that the agent has to avoid.
Should the agent propose an action that would cause a collision with the obstacle, it simply retains its original position.
An example of the environment is shown in Fig.~\ref{fig:framework_example}.
To successfully reach the goal position, the Euclidian distance of the position has to fulfill
\begin{equation}
    \lVert(x,y)^{agent} - (x,y)^{goal}\rVert \leq \epsilon.\label{eq:goal_reached}
\end{equation}

The agent can choose a continuous step size in $x$- and $y$- directions, limited to a maximum step size of 1 unit, resulting in the action space $\mathcal{A} = [-1, 1]^2$.

Combined with a sparse reward structure, where only reaching the goal is rewarded, the problem becomes difficult to solve for typical RL agents, as there is no reward signal for the initial random behaviour.
We use a sparse reward of 500 for reaching the goal state
\begin{equation}
    r_{sparse} = r_{reach\_goal} - 1 = \begin{cases}
        500 & \text{if the agent reaches the goal} \\
        -1 & \text{otherwise}
\end{cases}\label{eq:sparse}
\end{equation}
where the penalty of $-1$ per timestep encourages the agent to reach the goal as early as possible.
We also experiment with additional dense rewards, where we assign the following reward for moving closer to the goal:
\begin{equation}
    r_{distance} = 2 \cdot \left( \lVert(x,y)^{agent}_{t-1} - (x,y)^{goal}\rVert - \lVert(x,y)^{agent}_{t} - (x,y)^{goal}\rVert \right).
\end{equation}

We also penalize collisions with the obstacle, where we use a fixed value of $r_{obstacle\_collision} = -2$ per collision.
The overall dense reward is thus:
\begin{equation}
    r_{dense} = r_{reach\_goal} + r_{distance} + r_{obstacle\_collision} - 1\label{eq:dense}
\end{equation}


The observation space consists of the position of the agent, the goal position, and the two endpoints of the obstacle, if present.
Episodes have a horizon $T = 100$ and are terminated if $t = 100$ or the agent reaches the goal state according to \eqref{eq:goal_reached}.


\paragraph{Baseline}
For the experiments we use a baseline agent that moves directly towards the goal point, choosing the maximum step for each coordinate.
This behaviour is optimal for all cases without the obstacle blocking the path.
However, the baseline is not able to avoid the obstacle, meaning it will get stuck behind it at the last position before a collision.

\paragraph{Agent and shadow mode}
We use DDPG, an off-policy, actor-critic algorithm that uses a replay buffer to store transitions from which it learns the $Q$-function.
The implementation is built upon \textit{Stable Baselines 3}~\citep{stable-baselines}, with some modifications for calculating $Q(s_t, a_t^a)$ and $Q(s_t, a_t^b)$.
Some elements of the shadow mode, such as the truncation of $a_t^{decision}$ or the calculation of the regularization penalty~\eqref{eq:reg-action}, are handled by an environment wrapper.

\subsection{Results}
In this section we present all training results of the proposed approaches.
We evaluate the algorithms using 5 random seeds and report their mean performance on a test set throughout training, as well as their standard deviations.

\begin{figure}[ht]
    \centering
        \resizebox{!}{9em}{%
    \begin{tikzpicture}
        \begin{axis}[
            xlabel={training steps},
            ylabel={reward},
            ymin=-160, ymax=560,
            ytick={-100,0,100,200,300,400,500},
            legend pos=north west,
            ymajorgrids=true,
            grid style=dashed,
            skip coords between index={13}{20},
        ]

        \addplot[
            color=TUMBlue,
            mark=o,
            error bars/.cd, y fixed, y dir=both, y explicit
            ]
            table [x=timesteps1, y=results1, y error=results1_var, col sep=comma] {figures/data/agent-only/standard_002_sparse_jsrl.csv};
            \addlegendentry{sparse, $\epsilon = 0.02$}

        \addplot[
            color=TUMOrange,
            mark=square,
            error bars/.cd, y fixed, y dir=both, y explicit
            ]
            table [x=timesteps1, y=results1, y error=results1_var, col sep=comma] {figures/data/agent-only/standard_020_sparse_jsrl.csv};
            \addlegendentry{sparse, $\epsilon = 0.2$}

        \addplot[
            color=TUMGray,
            mark=triangle,
            error bars/.cd, y fixed, y dir=both, y explicit
            ]
            table [x=timesteps1, y=results1, y error=results1_var, col sep=comma] {figures/data/agent-only/standard_002_dense_jsrl.csv};
            \addlegendentry{dense, $\epsilon = 0.02$}

        \addplot+[
            color=TUMGreen,
            mark=diamond,
            error bars/.cd, y fixed, y dir=both, y explicit
            ]
            table [x=timesteps1, y=results1, y error=results1_var, col sep=comma] {figures/data/agent-only/standard_020_dense_jsrl.csv};
            \addlegendentry{dense, $\epsilon = 0.2$}

        \legend{}
        \end{axis}

        \end{tikzpicture}
        }
        \resizebox{!}{9em}{%
        \begin{tikzpicture}
        \begin{axis}[
            xlabel={training steps},
            ylabel={goal-reaching rate},
            ymin=-0.1, ymax=1.1,
            legend pos= outer north east,
            ymajorgrids=true,
            grid style=dashed,
            skip coords between index={13}{20},
        ]

        \addplot[
            color=TUMBlue,
            mark=o,
            error bars/.cd, y fixed, y dir=both, y explicit
            ]
            table [x=timesteps1, y=is_goal_reached1, y error=is_goal_reached1_var, col sep=comma] {figures/data/agent-only/standard_002_sparse_jsrl.csv};
            \addlegendentry{sparse, $\epsilon = 0.02$}

        \addplot[
            color=TUMOrange,
            mark=square,
            error bars/.cd, y fixed, y dir=both, y explicit
            ]
            table [x=timesteps1, y=is_goal_reached1, y error=is_goal_reached1_var, col sep=comma] {figures/data/agent-only/standard_020_sparse_jsrl.csv};
            \addlegendentry{sparse, $\epsilon = 0.2$}

        \addplot[
            color=TUMGray,
            mark=triangle,
            error bars/.cd, y fixed, y dir=both, y explicit
            ]
            table [x=timesteps1, y=is_goal_reached1, y error=is_goal_reached1_var, col sep=comma] {figures/data/agent-only/standard_002_dense_jsrl.csv};
            \addlegendentry{dense, $\epsilon = 0.02$}

        \addplot+[
            color=TUMGreen,
            mark=diamond,
            error bars/.cd, y fixed, y dir=both, y explicit
            ]
            table [x=timesteps1, y=is_goal_reached1, y error=is_goal_reached1_var, col sep=comma] {figures/data/agent-only/standard_020_dense_jsrl.csv};
            \addlegendentry{dense, $\epsilon = 0.2$}

        \end{axis}

    \end{tikzpicture}
        }
    \caption{\small Reward and goal reaching rate for training with DDPG without shadow mode}
    \label{fig:results-agent-only}
\end{figure}
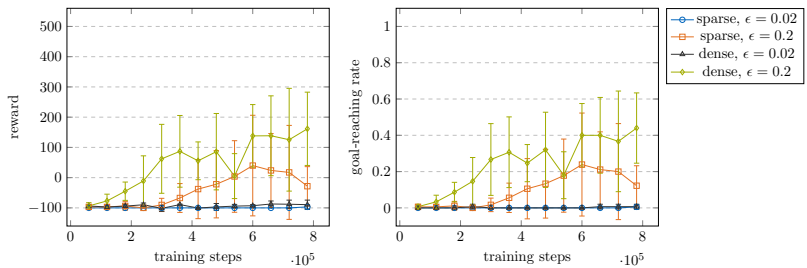

Finding a good policy without the help of the baseline is very difficult.
Fig.~\ref{fig:results-agent-only} shows that the agent fails to learn entirely when not using the shadow mode, as it does not receive any signal if the reward is sparse.
If we use a dense reward, the agent does move toward the goal, however, it still fails to precisely move to the goal position, especially if $\epsilon$ is small. 

\definecolorset{RGB}{TUM}{}{%
	Black,   0,   0,   0;%
	Gray,   51,  51,  51;%
	White, 255, 255, 255;%
	Blue, 0, 101, 189;%
	Ivory,  218, 215, 203;%
	Orange, 227, 114,  34;%
	Green,  162, 173,   0%
}

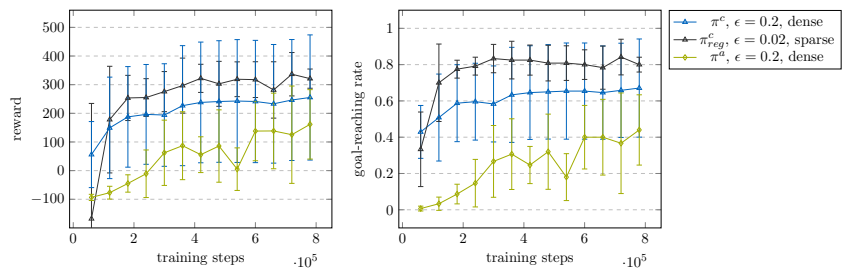
\begin{figure}[ht]
    \centering
        \resizebox{!}{9em}{%
    \begin{tikzpicture}
        \begin{axis}[
            xlabel={training steps},
            ylabel={reward},
            ymin=-200, ymax=560,
            ytick={-100,0,100,200,300,400,500},
            ymajorgrids=true,
            grid style=dashed,
            skip coords between index={13}{20},
        ]

        \addplot[
            color=TUMGray,
            mark=triangle,
            error bars/.cd, y fixed, y dir=both, y explicit
            ]
            table [x=timesteps1, y=results3, y error=results3_var, col sep=comma]
                {figures/data/agent-decision/agent_decision_002_sparse_jsrl_adp.csv};
            \addlegendentry{$\pi^c$, $\epsilon = 0.02$, sparse reward}
            
        \addplot[
            color=TUMBlue,
            mark=triangle,
            error bars/.cd, y fixed, y dir=both, y explicit
            ]
            table [x=timesteps1, y=results3, y error=results3_var, col sep=comma]
                {figures/data/agent-decision/agent_decision_020_dense_jsrl.csv};
            \addlegendentry{$\pi^c$, $\epsilon = 0.2$, dense reward}

        \addplot+[
            color=TUMGreen,
            mark=diamond,
            error bars/.cd, y fixed, y dir=both, y explicit
            ]
            table [x=timesteps1, y=results1, y error=results1_var, col sep=comma] {figures/data/agent-only/standard_020_dense_jsrl.csv};
            \addlegendentry{$\pi^a$, $\epsilon = 0.2$, dense}

        \legend{}

        \end{axis}

        \end{tikzpicture}
        }
        \resizebox{!}{9em}{%
        \begin{tikzpicture}
        \begin{axis}[
            xlabel={training steps},
            ylabel={goal-reaching rate},
            ymin=-0.1, ymax=1.1,
            legend pos= outer north east,
            ymajorgrids=true,
            grid style=dashed,
            skip coords between index={13}{20},
        ]

        \addplot[
            color=TUMBlue,
            mark=triangle,
            error bars/.cd, y fixed, y dir=both, y explicit
            ]
            table [x=timesteps1, y=is_goal_reached3, y error=is_goal_reached3_var, col sep=comma]
                {figures/data/agent-decision/agent_decision_020_dense_jsrl.csv};
            \addlegendentry{$\pi^c$, $\epsilon = 0.2$, dense}

        \addplot[
            color=TUMGray,
            mark=triangle,
            error bars/.cd, y fixed, y dir=both, y explicit
            ]
            table [x=timesteps1, y=is_goal_reached3, y error=is_goal_reached3_var, col sep=comma]
                {figures/data/agent-decision/agent_decision_002_sparse_jsrl_adp.csv};
            \addlegendentry{$\pi^c_{reg}$, $\epsilon = 0.02$, sparse}

        \addplot+[
            color=TUMGreen,
            mark=diamond,
            error bars/.cd, y fixed, y dir=both, y explicit
            ]
            table [x=timesteps1, y=is_goal_reached1, y error=is_goal_reached1_var, col sep=comma] {figures/data/agent-only/standard_020_dense_jsrl.csv};
            \addlegendentry{$\pi^a$, $\epsilon = 0.2$, dense}
            
        \end{axis}

    \end{tikzpicture}
        }

    \caption{\small Test Reward and goal reaching rate for training with DDPG in shadow mode, $\pi^c_{reg}$ regularized with regularization penalty~\eqref{eq:reg-action} with strength $\lambda = 2$, and control authority by agent. Both combined policies outperform the RL agent's policy $\pi^a$ that was not trained in shadow mode, even when it is trained with a dense reward and $\epsilon = 0.2$.}
    \label{fig:results-agent-decision-comp}
\end{figure}

When training in shadow mode using an additional agent action as in Sec.~\ref{subsubsec:agent_decision} and a sparse reward, the RL agent is prone to exploiting the baseline without searching for better strategies.
The combined strategy thus reduces to $\pi^b$ and provides no benefit.
If we introduce the dense reward~\eqref{eq:dense}, we learn a combined policy that strongly outperforms the baseline policy (see Fig.~\ref{fig:results-agent-decision-comp}).
The resulting combined policy $\pi^c$ relies heavily on the baseline, using its action in over 90 \% of the steps but outperforms $\pi^b$ as desired.

\definecolorset{RGB}{TUM}{}{%
	Black,   0,   0,   0;%
	Gray,   51,  51,  51;%
	White, 255, 255, 255;%
	Blue, 0, 101, 189;%
	Ivory,  218, 215, 203;%
	Orange, 227, 114,  34;%
	Green,  162, 173,   0%
}

We find that regularization according to~\eqref{eq:reg-action} can greatly improve the performance of the combined policy, and reduces variances in the reward (see Fig.~\ref{fig:results-agent-decision-comp}).
With regularization, we can even train on sparse rewards and with the smaller $\epsilon = 0.02$.

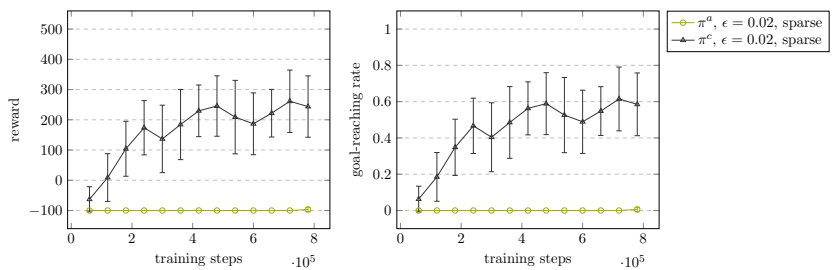
\begin{figure}[ht]
    \centering
        \resizebox{!}{9em}{%
    \begin{tikzpicture}
        \begin{axis}[
            xlabel={training steps},
            ylabel={reward},
            ymin=-160, ymax=560,
            ytick={-100,0,100,200,300,400,500},
            ymajorgrids=true,
            grid style=dashed,
            skip coords between index={13}{20},
        ]

        \addplot[
            color=TUMGreen,
            mark=o,
            error bars/.cd, y fixed, y dir=both, y explicit
            ]
            table [x=timesteps1, y=results1, y error=results1_var, col sep=comma] {figures/data/agent-only/standard_002_sparse_jsrl.csv};
            \addlegendentry{$\pi^a$, $\epsilon = 0.02$, sparse}

        \addplot[
            color=TUMGray,
            mark=triangle,
            error bars/.cd, y fixed, y dir=both, y explicit
            ]
            table [x=timesteps1, y=results3, y error=results3_var, col sep=comma]
                {figures/data/max-q-val/max_q_002_sparse_jsrl.csv};
            \addlegendentry{$\pi^c$, $\epsilon = 0.2$}

        \legend{}
        \end{axis}

        \end{tikzpicture}
        }
        \resizebox{!}{9em}{%
        \begin{tikzpicture}
        \begin{axis}[
            xlabel={training steps},
            ylabel={goal-reaching rate},
            ymin=-0.1, ymax=1.1,
            legend pos= outer north east,
            ymajorgrids=true,
            grid style=dashed,
            skip coords between index={13}{20},
        ]

        \addplot[
            color=TUMGreen,
            mark=o,
            error bars/.cd, y fixed, y dir=both, y explicit
            ]
            table [x=timesteps1, y=is_goal_reached1, y error=is_goal_reached1_var, col sep=comma] {figures/data/agent-only/standard_002_sparse_jsrl.csv};
            \addlegendentry{$\pi^a$, $\epsilon = 0.02$, sparse}

        \addplot[
            color=TUMGray,
            mark=triangle,
            error bars/.cd, y fixed, y dir=both, y explicit
            ]
            table [x=timesteps1, y=is_goal_reached3, y error=is_goal_reached3_var, col sep=comma]
                {figures/data/max-q-val/max_q_002_sparse_jsrl.csv};
            \addlegendentry{$\pi^c$, $\epsilon = 0.02$, sparse}

        \end{axis}

    \end{tikzpicture}
        }

    \caption{\small Reward (sparse) and goal reaching rate for training with DDPG in shadow mode and control authority by $Q$-value compared to standard training}
    \label{fig:results-max-q-val}
\end{figure}

We are also able to train with sparse rewards using the $Q$-value as described in Sec.~\ref{subsubsec:max_q_val},  as shown in Fig.~\ref{fig:results-max-q-val}. 
This approach also leads to a combined policy which chooses the baseline action in over 90 \% of steps.
However, more of the suggested agent actions are used during early training than with the agent taking control authority.
This is due to the random initialization of the $Q$-function, leading to less exploitation of the baseline and more exploration initially.

\definecolorset{RGB}{TUM}{}{%
	Black,   0,   0,   0;%
	Gray,   51,  51,  51;%
	White, 255, 255, 255;%
	Blue, 0, 101, 189;%
	Ivory,  218, 215, 203;%
	Orange, 227, 114,  34;%
	Green,  162, 173,   0%
}




Control authority by $Q$-value does not require an additional hyperparameter (such as regularization strength $\lambda$), but does not quite reach the level of a decision by the agent itself in our experiments.
All in all, control authority by agent or $Q$-value are both viable options for training in shadow mode and generate performant combined policies.

\begin{figure}
    \centering
    \includegraphics[scale=0.27]{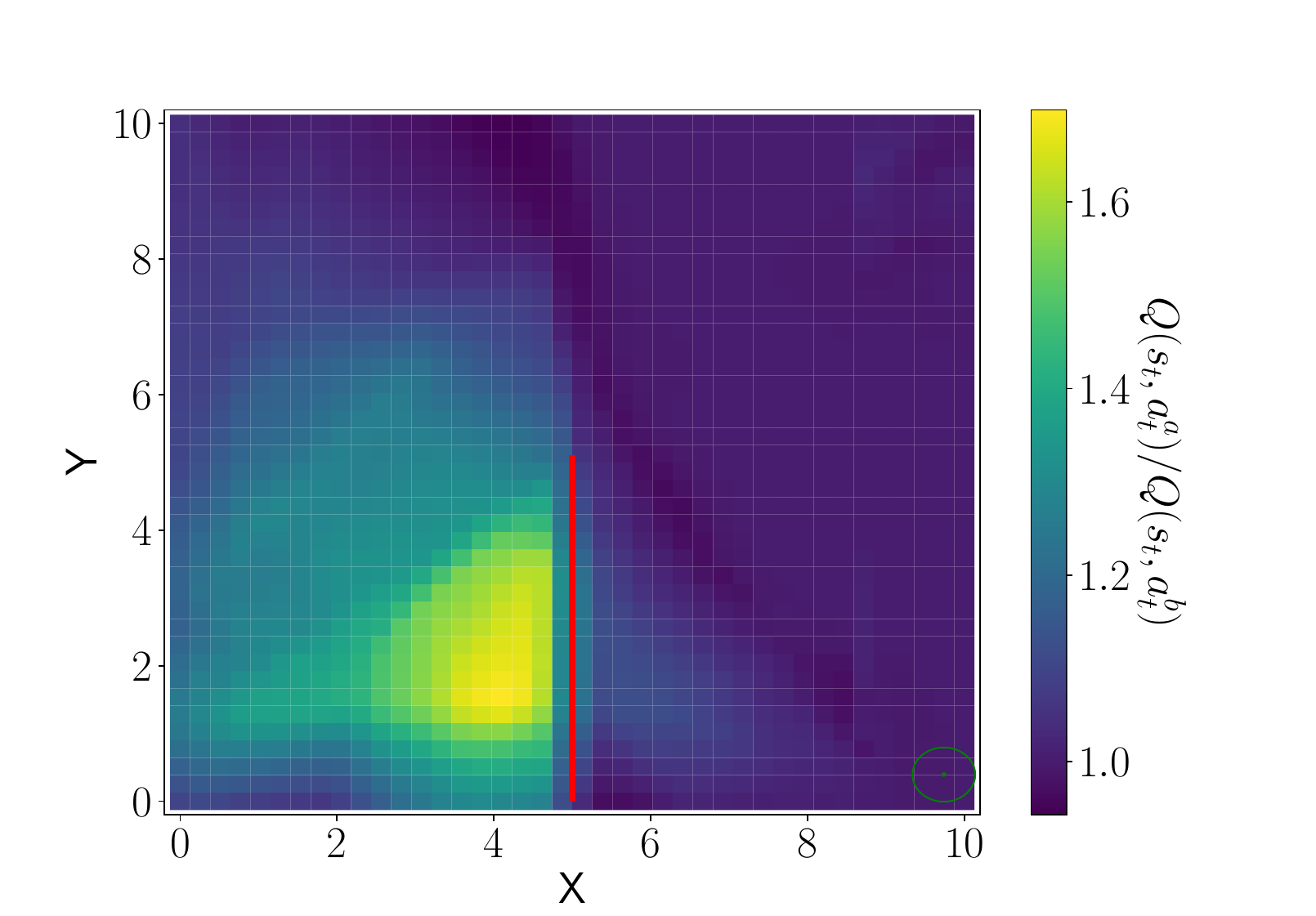}
\caption[Uncertainty distribution]{\small The figure contains a heatmap of the decision criterion $Q(s_t,a_t^a)/Q(s_t,a_t^b)$ for all possible positions of the agent with the given obstalce (red bar) and the target (green dot, $\epsilon = 0.02$, with green circle added for visibility). The baseline is acting for all values smaller or equal to 1 and the agent in all other regions.}\label{fig:distribution}
\end{figure}

An exemplary distribution of the decision over all possible agent positions is given in Fig.~\ref{fig:distribution}.
The agent's action is primarily favoured behind the obstacle, which is exactly the desired behaviour.
When the obstacle is evaded, the baseline can take over again.

\section{Conclusion} \label{sec:conclusions}
We present a novel approach to train a reinforcement learning agent in a so-called shadow mode, which enables training on real systems while still efficiently operating that system.
The proposed method of training enables us to combine powerful RL techniques with existing controllers and simultaneously leverages the guidance that an existing policy provides for the training of an RL agent.
We present two approaches for switching between the RL agent and the baseline, and demonstrate their efficacy on a reach-avoid task.
We show that the RL agent successfully identifies a performant policy and automatically takes control of the system.
Thus, the algorithm can readily be deployed on a wide variety of systems.


\bibliography{main}

\end{document}